\documentclass[fleqn,10pt]{wlscirep}
\usepackage[utf8]{inputenc}
\usepackage[T1]{fontenc}
\usepackage{moreverb,url}
\usepackage{graphicx}
\usepackage{times}
\usepackage{amsmath}
\usepackage{amssymb}
\usepackage{url}
\usepackage{ctable}
\usepackage{multirow}
\usepackage{stfloats}
\usepackage{array}
\usepackage{gensymb}
\usepackage{balance}
\usepackage{float}
\usepackage{caption}
\usepackage{subcaption}
\usepackage{lineno}
\usepackage{adjustbox}
\title{Analysis and Evaluation of Explainable Artificial Intelligence on Suicide Risk Assessment }

\author[1]{Hao Tang}
\author[1]{Aref Miri Rekavandi}
\author[2,3,4,*]{Dharjinder Rooprai}
\author[5,6]{Girish Dwivedi}
\author[7]{Frank Sanfilippo}
\author[8]{Farid Boussaid}
\author[1,*]{Mohammed Bennamoun}
\affil[1]{Department of Computer Science $\&$ Software Engineering, The University of Western Australia, Perth Australia}
\affil[2]{Consultant Psychiatrist, Armadale Mental Health Service, Perth Australia}
\affil[3]{Director of Medical Services, Bethesda Clinic, Perth Australia}
\affil[4]{Clinical Lead, Co-HIVE Mental Health, Australia}
\affil[5]{Advanced clinical and translational cardiovascular imaging, Harry Perkins Institute of Medical Research, The University of Western Australia, Perth Australia}
\affil[6]{Department of Cardiology,  Fiona Stanley Hospital, WA, Australia}
\affil[7]{School of Population and Global Health, University of Western Australia, Perth Australia}
\affil[8]{Department of Electrical, Electronic and Computer Engineering, The University of Western Australia, Perth Australia}

\affil[*]{Corresponding Authors. Emails: mohammed.bennamoun@uwa.edu.au, dharjinder.rooprai@health.wa.gov.au }

\begin{abstract}
This study investigates the effectiveness of Explainable Artificial Intelligence (XAI) techniques in
predicting suicide risks and identifying the dominant causes for such behaviours.  Data augmentation techniques and ML models are utilized to predict the associated risk. Furthermore, SHapley Additive exPlanations (SHAP) and correlation analysis are used to rank the importance of variables in predictions. Experimental results indicate that Decision Tree (DT), Random Forest (RF) and eXtreme Gradient Boosting (XGBoost) models achieve the best results while DT has the best performance with an accuracy of $95.23\%$ and an Area Under Curve (AUC) of 0.95.
As per SHAP results, anger problems, depression, and social isolation are the leading variables in predicting the risk of
suicide, and patients with good incomes, respected occupations, and university education have the least risk. Results demonstrate the effectiveness of machine learning and XAI framework for suicide risk prediction, and they can assist psychiatrists in understanding complex human behaviours and can also assist in reliable clinical decision-making.
\end{abstract}
\begin{document}

\flushbottom
\maketitle
%
%
\thispagestyle{empty}

\section*{Introduction}

Suicide accounted for $1.3\%$ of global deaths and was the $17^{th}$ leading cause of death in 2019. With more than 700,000 people committing suicide yearly, $77\%$ of suicidal behaviours occurred in developing countries \cite{world2021suicide}. An American survey also shows that $93\%$ of adults believe that suicides can be delayed or prevented if psychiatrists intervene effectively and immediately. According to World Health Organisation’s statistics, suicidal adults may have attempted 20 times before their death \cite{american2021suicide}.  According to a report published by the Centres for Disease Control and Prevention, middle-aged white men have the highest suicidal risk in America \cite{american2021suicide}, and suicide was the leading cause of death among Australian teenagers aged 15 to 24 in 2019 \cite{australian2021suicide}. The current tools and solutions for suicide prevention mostly rely on self-reported measures, such as questionnaires and interviews, which can be subjective or multimodal data \cite{gao2023multimodal,kamimura2022associations} which is not easy to collect. Furthermore, traditional clinical risk assessment tools have been shown to be not sufficiently accurate to identify moderate-and high-risk patients \cite{su2020machine}. Two recent systematic reviews \cite{carter2017predicting,belsher2019prediction} have evaluated various scales to predict the risk of suicide but have found overall low Positive Predictive Value (PPV). Hence, there is a critical need to develop technologies and models that can assist psychiatrists and mental health professionals to accurately stratify risk, enable precision medicine, and allocate resources.\\
Over the past decade, researchers have proposed various Machine Learning (ML) solutions and frameworks to enhance the performance of suicide prediction; however, since they are primarily ``black box'' units and not interpretable, it is challenging to use them in clinical treatments. The objectives of this study are threefold. \textbf{First}, we review the related works to summarise the existing ML models used for suicide prediction. \textbf{Second}, we select and integrate suitable ML algorithms and use data augmentation methods to assess the feasibility of ML models for suicide prediction. \textbf{Finally}, we determine which variables contribute the most, using an Explainable Artificial Intelligence (XAI) framework to determine features' importance and visualize the underlying logic behind the predictions. ML is a branch of computer science that uses historical data to train models and make predictions about future trends through building models, testing and improving models. In recent years, there has been a rapid growth and progress in the field of computer science, including ML, Computer Vision, Artificial Intelligence (AI), Natural Language Processing (NLP) which has led to the development of new tools and techniques to predict the risk of physical and psychological illnesses \cite{bejan2022improving}. For instance, these technologies have been implemented to predict the possibility of heart attacks \cite{aghamohammadi2019predicting}, liver diseases \cite{choudhary2021efficient,li2022predictors}, alcohol disorders \cite{benjet2022risk}, human emotion disorders \cite{chen2020comments, bendjoudi2020audio}, depressions \cite{zulfiker2021depth} etc. In the past decade, studies have also shown that ML can be effective in predicting suicide risk prediction \cite{linthicum2019machine,bernert2020artificial}.\\
In recent years, numerous research studies have used ML techniques to predict suicide. For example \cite{wang2021learning}
integrated a C-Attention Network architecture with multiple ML models to identify individuals at risk for  suicide. The three-stage suicide theory and prior emotions work were also introduced to examine suicidal thoughts. In the sub-task of predicting suicide attempts in 30 days, traditional ML models had superior  performance compared to the baseline in predicting suicide attempts, with an F1 score of $0.741$  and an F2 score of $0.833$ (higher F-score shows better performance \cite{sokolova2006beyond}). Moreover, when predicting suicide within a six-month period, the C-Attention method also outperformed the baseline, achieving an F1 score of $0.737$  and  an F2 score of $0.833$. Other research has also utilized smartphone applications to gather data on outpatients' therapy and apply NLP techniques to assess patients’ suicide risk levels \cite{cohen2020feasibility}. The results showed that the Support Vector Machine (SVM) and Logistic Regression produced satisfactory calcification scores, while the extreme gradient model achieved the highest AUC value (0.78). The authors in \cite{cohen2020feasibility} highlighted the importance of using XAI tools to address the lack of explainability in traditional ML models, as it is crucial for psychiatrists to trust and rely on ML models. 
Similarly in \cite{miche2020prospective}, the authors compared the performance of four traditional models, namely logistic regression, Lasso, Ridge, and Random Forrest, using the epidemiological Early Developmental Stages of Psychopathology (EDSP) dataset. After conducting nested 10-fold cross-validation, they found that these models performed almost the same in terms of mean AUC values ranging from  0.824 to 0.829.  Furthermore, the RF model achieved the highest PPV of  $87\%$, which was significantly better than other models. In suicide prediction research, various types of surveys, questionnaires, and scales have been used. For instance, in \cite{ryu2019detection} the Korea National Health $\&$ Nutrition Examination Survey (KNHANES) and the Synthetic Minority Over-sampling TEchnique (SMOTE) were used to select patients with suicidal thoughts and to construct the dataset by resampling. After pre-processing, a Random Forrest (RF) algorithm was applied and the experimental results verified the feasibility of such techniques on the general population. The RF model achieved an AUC of 0.947 and an $88.9\%$ accuracy. Notably, the feature selection process identified days of feeling sick or in discomfort, daily smoking amount, and household composition as the most significant features that contributed to the prediction. \\
Traditional mathematical techniques produced less accurate results due to the complexity of input/output relationships in human behaviours. In \cite{kim2021phq} the authors used the Patient Health Questionnaire-9 (PHQ-9) to collect data from college students and used the Mini-International Neuropsychiatric Interview suicidality module to evaluate their suicide ideation. They applied ML models, including K-Nearest Neighbourhood (KNN), Linear Discriminant Analysis (LDA), and RF. Their results showed that the RF model had the best performance, with an AUC value of 0.841 and an accuracy of $94.3\%$. The positive and negative predictive values of the RF were also noteworthy, with values of  $84.95\%$ and $95.54\%$, respectively.
RF modes were also used in other research studies, such as  in \cite{shen2020detecting} to predict suicidal attempts on a self-report dataset collected from 4,882 Chinese medical students. The dataset included clinical features from multiple psychiatric scales, including the Self-rating Anxiety Scale (SAS), the Self-rating Depression Scale (SDS), the Epworth Sleepiness Scale (ESS), the Self-Esteem Scale (SES), and the Chinese version of Connor Davidson Resilience Scale (CD-RISC). After applying five-fold cross-validation to the model, the experimental results showed that the RF model achieved significant performance, with an AUC value of  0.925 and an accuracy of $90.1\%$ in suicide prediction. This study also made several noteworthy discoveries. For instance, it found that relationships with parents were among the top five predictors of college students' suicide risk, and participants with low care from fathers were associated with a higher risk of suicide.
ML algorithms have demonstrated potential in analyzing datasets from psychometric scales, such as the Suicide Crisis Inventory (SCI) and Columbia Suicide Severity Rating Scale (CSSRS) \cite{parghi2021assessing}. In order to improve model performance, the researchers 
 employed Gradient Boosting (GB) techniques to minimize prediction error and used SMOTE to generate artificial/synthetic data points. Their experimental results revealed that RF and GB algorithms performed the best, with precision values of $98.0\%$ and $94\%$ respectively for detecting short-term suicidal behaviours.
An artificial neural network classifier with 31 psychiatric scales and 10 sociodemographic elements was proposed to predict suicide and assess the performance of ML models as well as identify the most significant variables \cite{oh2017classification}. The classifier's accuracy for predicting suicide within one month, one year, and the whole lifetime were $93.7\%$, $90.8\%$, and $87.4\%$, respectively. In terms of the AUC, the highest was in one-month detection (0.93), followed by lifetime prediction (0.89) and 1-year (0.87). In their study, the Emotion Regulation Questionnaire (ERQ) has the highest impact, followed by the Anger Rumination Scale (ARS) and the Satisfaction With Life Scale (SWLS) \cite{oh2017classification}.\\
All the studies previously mentioned were designed to apply standard machine learning techniques to predict suicide risk using their own private and imbalanced dataset with a small number of records. Additionally, to score the importance of each variable, conventional correlation analysis was commonly used. In contrast, in this current work, we not only use conventional tools, but also employ data augmentation and state-of-the-art AI frameworks to effectively analyze and interpret data.
\section*{Results}

\begin{figure} \centering
		\includegraphics[width=0.7\linewidth]{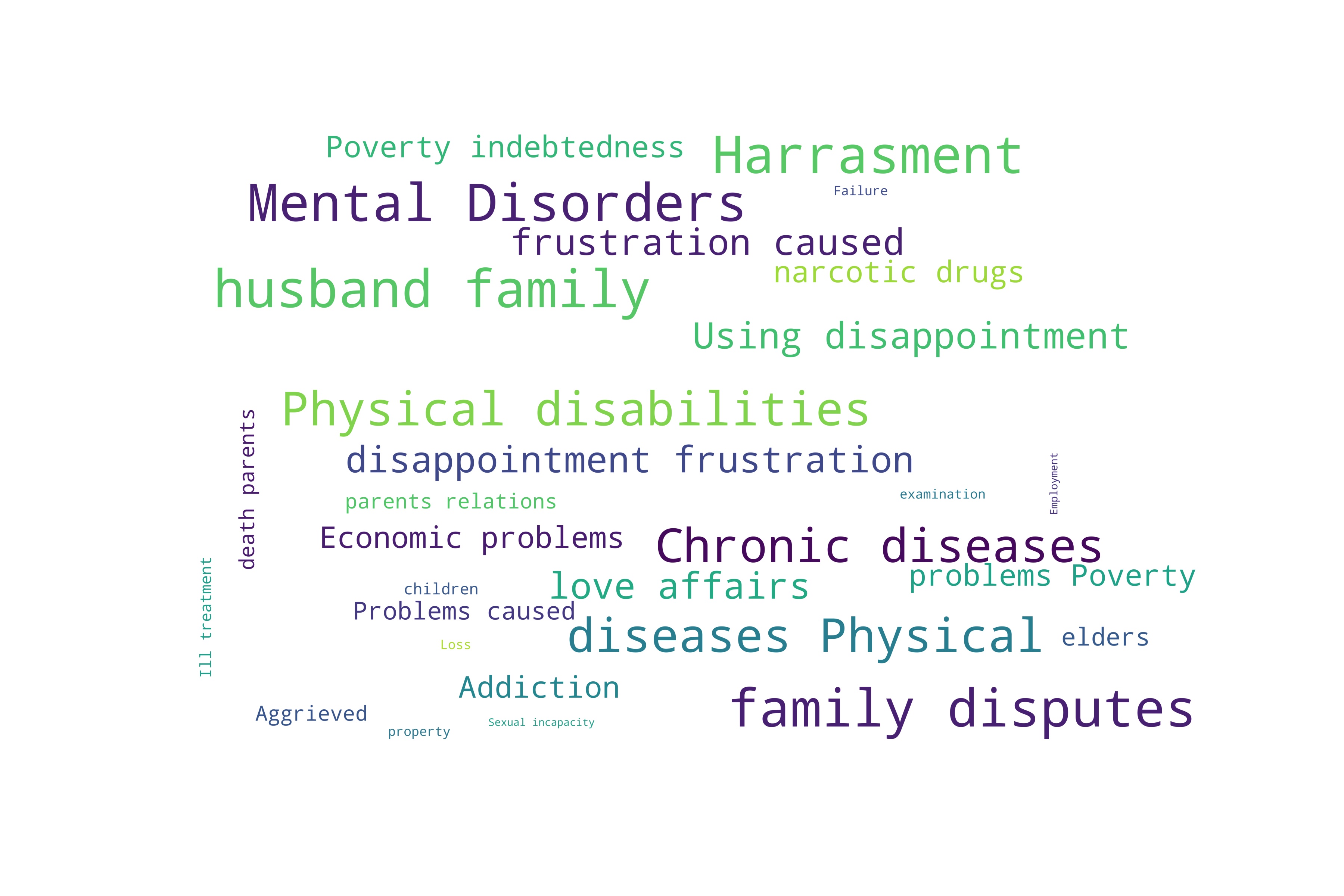}
		\caption{The word cloud of death reasons in the dataset. As shown, according to investigated data, physical and
psychiatric disorders such as ``physical disabilities,'' ``mental disorders,'' ``chronic diseases,'' and ``family disputes'' are the most common reasons for death.  }
	\label{word}
\end{figure}

\subsection*{Data Visualization}
\begin{figure*} \centering
		\includegraphics[width=0.7\linewidth]{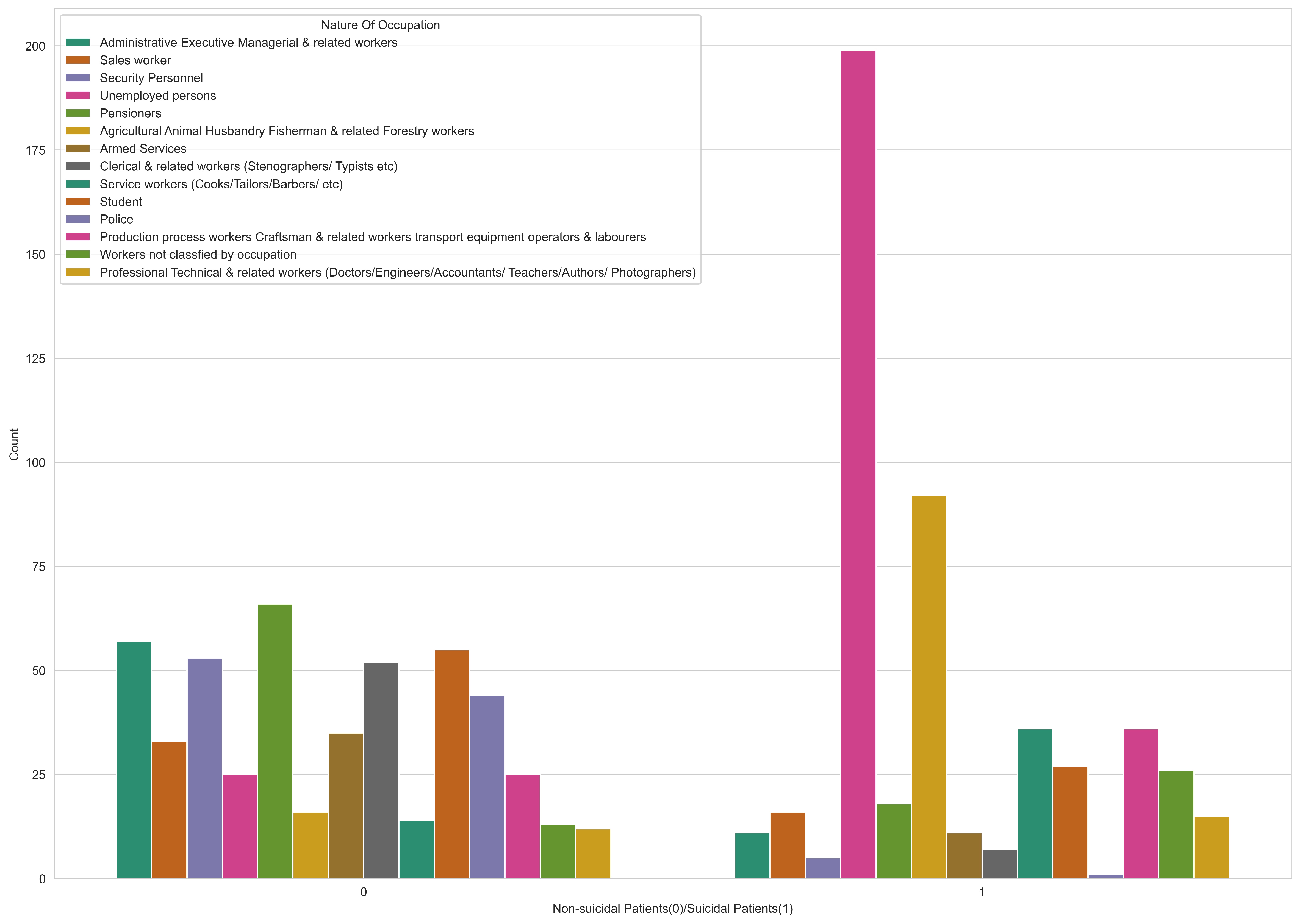}
		\caption{Counts of suicidal and non-suicidal patients among different occupations. Unemployment and agriculture
and forest-related workers are among high-risk patients and on the other hand, police officers and security personnels have the minimum risk. }
	\label{occ}
\end{figure*}

The word cloud, a popular method in the NLP area, provides an intuitive illustration of the word frequency for individuals to understand which words appear most frequently in the dataset.
In the word cloud illustrated in Fig. \ref{word}, we observe that among the reasons for death, there are some high-frequency words, including ``physical disabilities,'' ``mental disorders,'' ``chronic diseases,'' and ``family disputes.'' We can assume that many suicidal patients also suffer from these physical and psychiatric disorders. Figure \ref{occ} visualizes the count of suicidal and non-suicidal patients in each occupation. Unemployment is the leading reason for suicidal behaviours, while agriculture and forest-related workers have a higher risk of committing suicide. An interesting discovery is that few police officers commit suicide, and the suicidal rate of administrative managers is relatively low among occupations. The conclusion of our visualization is similar to the results in previous studies \cite{goldstein2015peeking}, which suggests that patients with a good life income and people who can gain respect from their occupations are less prone to committing suicide.

\begin{table}[t!]
\caption{Models performances (in $\%$) in predicting suicide behaviours. We observe that DT and RF are performing the best in identifying patients with a high risk of suicidal acts.}
  \centering
 \begin{adjustbox}{scale=0.77,angle=0}
    \begin{tabular}{llllll}
    \hline
   \textbf{Models $\downarrow$/Metric$\rightarrow$}& Accuracy & Precision & Recall & F1 Score & AUC \\
    \hline
    SVM	& 87.46	&87.77&	86.19&86.97&	87\\
    LR&	88.63&	88.95&	87.83&	88.39&	89\\
    DT&	\textbf{95.23}&	\textbf{96.98} &93.22&	\textbf{95.07}&	\textbf{95}\\
    RF&	95.20&	96.65&\textbf{93.28}&	94.93&	\textbf{95}\\
    Linear SVC &	87.51&	87.45&	86.52&	86.98&	89\\
    Perceptron (iter=10)&	89.35&	89.17&	88.68&	88.93&	89\\
    XGBoost&	94.56&	95.70&	92.65&	94.15&	94\\
    
    \bottomrule
    \end{tabular}%
    \end{adjustbox}
  \label{performance}%
\end{table}%
\begin{figure} \centering
		\includegraphics[width=0.65\linewidth]{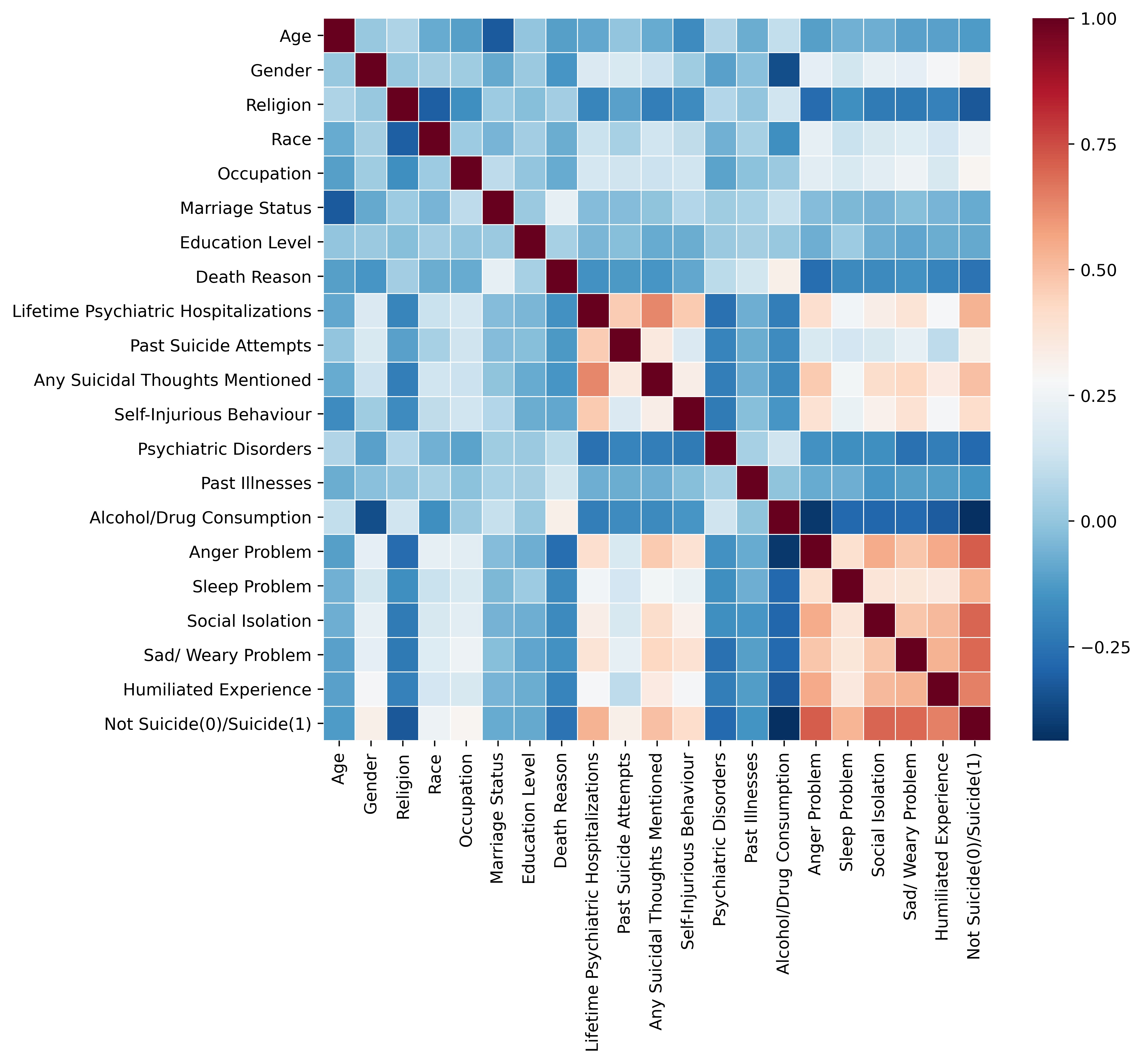}
		\caption{The correlation matrix of suicide-related variables. Results show a strong correlation between suicidal acts and anger problems, sleep problems, social isolation, depression
problems, humiliating experiences, past suicide attempts,
suicidal thoughts, self-injuries, and psychiatric disorders. }
	\label{corr}
\end{figure}
\subsection*{ML Performance}
This section reports the performance of our ML algorithms. We have implemented Random Forrest (RF), Decision Tree (DT), Logistic Regression (LR), Support Vector Machine (SVM), linear Support Vector Classification (SVC), Perceptron, and eXtreme Gradient Boosting (XGBoost) to predict the risk of suicide based on patients’ records. To avoid the influence of contingency on experimental results, we tried different sample sizes, train-test splitting percentages, and random seeds. We collected performance statistics and calculated their average performance. Table \ref{performance} lists the performance of each model under different evaluation indicators.
According to the above statistics, DT, RF, and XGBoost have excellent performances among the seven ML models, with AUC values higher than 0.94. In terms of the other evaluation metrics, DT achieves the highest accuracy, precision, and F1 score indicators, which are respectively $95.23\%$, $96.98\%$, and $95\%$. RF has the best recall performance, which is $93.28\%$. To enhance the credibility of our models and understand the underlying reasons for their high performance, this research introduces a correlation matrix and XAI model to further investigate the reasons behind the above performances.

\subsection*{Correlation Analysis}
In this section, we use the correlation function in Seaborn library and the heat-map function to analyze the correlation among attributes in the dataset.   Considering that most variables in our dataset are categorical and non-continuous variables, we use the Spearman correlation to perform the analysis.
Figure \ref{corr} clearly illustrates the correlation between every two variables. According to the colour bar on the right-hand side, when the correlation colour between two variables is closer to 1, it is coded with dark red colour showing a significant positive correlation. A red area in the bottom right corner of the figure indicates that these variables are highly related. The heat-map shows a strong correlation between suicide and anger problems, sleep problems, social isolation, depression problems, and humiliating experiences.
Moreover, the light red area in the center demonstrates a moderate correlation between the patient's suicidal risk with past suicide attempts, suicidal thoughts, self-injuries, and psychiatric disorders. The above analysis proves that every single variable, which mostly measures mental issues, can considerably contribute to the model prediction and the model would become more powerful when we combine all of these variables for prediction.

\subsection*{Analysis by Explainable AI}
\begin{figure*} \centering
		\includegraphics[width=0.95\linewidth]{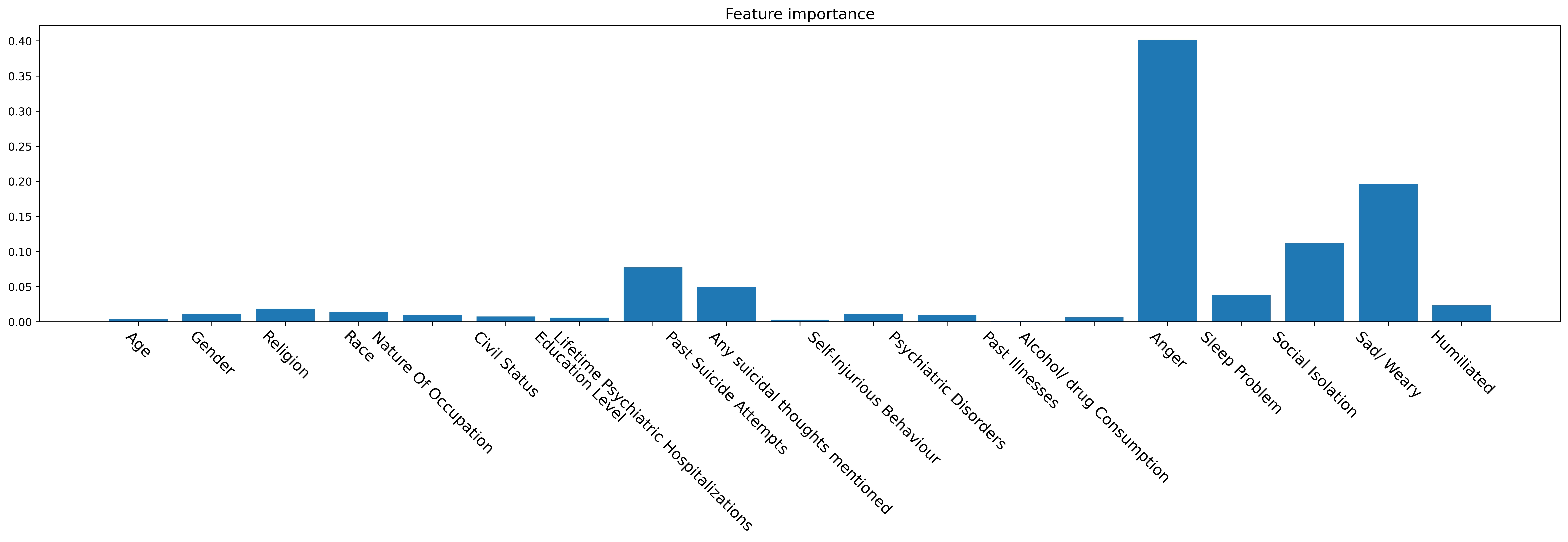}
		\caption{Traditional feature importance analysis provided by XGBoost predicting model. Similar to previous results, anger problem is the most important variable in suicidal risk prediction. }
	\label{python}
\end{figure*}

With the growing need to understand the underlying logic of ML models, studies have introduced the XAI framework to analyze the contribution of variables in model prediction. The generalization of SHapley Additive exPlanations (SHAP) and local interpretable model-agnostic explanations methods extends the use of XAI in the ML domain. Python package XGBoost provides library functions to calculate the importance of features that contribute to the final model. Figure \ref{python} demonstrates the features' importance in predicting suicide. It is evident that anger problem is the dominant variable correlated with suicidal behaviours. Mental health issues, including depression problems, social isolation, sleeping problem, and humiliating experiences are in next places and need psychiatrists' attention. Meanwhile, past suicidal attempts and suicidal ideation are important factors for a patient who commits suicide.
Some nonlinear models, such as XGBoost, have significantly stronger prediction accuracy. However, their characteristics also make their interpretability inferior to linear models, which impedes them from being promoted in practical clinical diagnosis. The Shapley value is a calculation method for fair distribution in cooperative game theory \cite{lipovetsky2021game}. SHAP is an additive interpretation model based on Shapley's value. The model produces a prediction value for each prediction sample, and the SHAP value is the value assigned to each feature in the sample \cite{lundberg2017unified}. Suppose the $i^{th}$  sample is $x_i$, the $j^{th}$ feature of the $i^{th}$  sample is $x_{i,j}$, the model predicted sample of the $i^{th}$  sample is $y_i$, the baseline of the entire model (usually the mean value of the target variable of all samples) is $y_{base}$, then the SHAP value is given by 
\begin{equation}\label{eq2}
    y_i=y_{base}+f(x_{i,1})+f(x_{i,2})+\cdots+f(x_{i,k}),
\end{equation}
where $f(x_{i,j})$ is the SHAP value of $x_{i,j}$, and $k$ is the total number of variables. Intuitively, $f(x_{i,j})$ is the amount of contribution in forming $y_i$. When $f(x_i,j)>0$, the feature improves the predicted value and has a positive effect. In contrast, negative values mean that these features decrease the predicted value. Compared to the traditional feature importance method, the advantage of SHAP is that it can reflect the importance of variable values in each sample, and it also shows the positive and negative contributions of variables. This can help to assign a contribution share to each variable, where high positive values and high negative values show strong direct and reverse correlation with the output predicted risk, showing which variables act as a driver to commit suicide and which variables prevent it. On the other hand, SHAP values around zero show the irrelevancy of the variable with output.

\begin{figure*} \centering
		\includegraphics[width=1\linewidth]{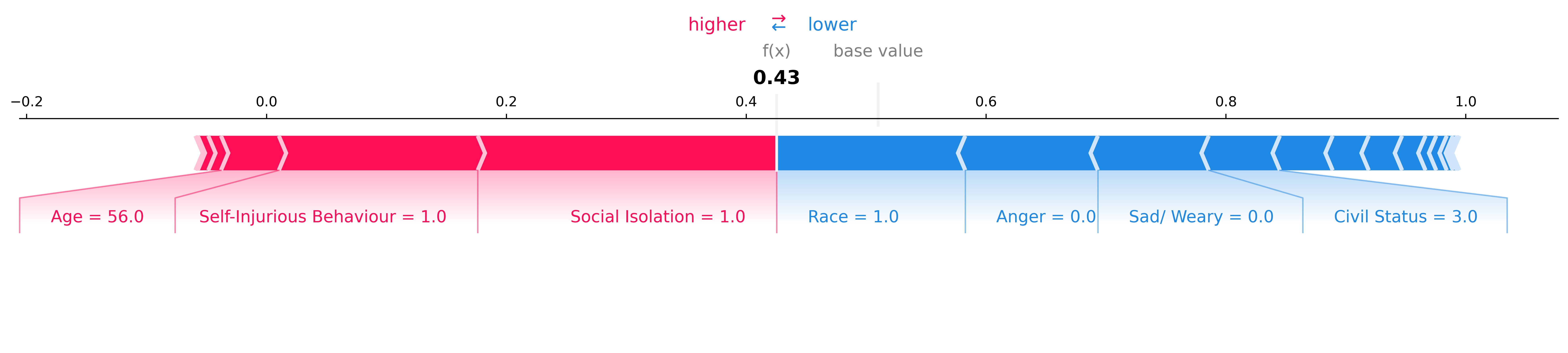}
		\caption{Variables with positive and negative contributions using SHAP analysis for a random sample. For this particular sample, the model predicts that this person is at risk of suicide commitment because of his/her age,  past self-injury experiences, and social isolation problems. }
	\label{shap}
\end{figure*}

\begin{table}[t!]
\caption{SHAP value of a single sample.}
  \centering
 \begin{adjustbox}{scale=0.7,angle=0}
    \begin{tabular}{llll}
    \hline
   Feature ID & Feature & Feature Value & SHAP \\
    \hline
    0&	Age&	56&	0.04823\\
    1&	Gender&	1&	0.0019\\
    2&	Religion&	3	&-0.03007\\
    3&	Race&	1&	-0.15723\\
    4&	Nature of occupation&	5&	-0.04402\\
    5&	Civil status&	3&	-0.05929\\
    6&	Education level&	0&	-0.00184\\
    7&	Lifetime psychiatric hospitalisations&	0	&-0.01955\\
    8&	Past suicide attempts&	0	&-0.00517\\
    9&	Any suicidal thoughts mentioned&	0	&-0.00878\\
    10&	Self-injurious behaviour&	1	&0.16583\\
    11&	Psychiatric disorders&	3	&-0.00353\\
    12&	Past illnesses&	4	&0.00875\\
    13&	Alcohol/ drug consumption&	2&	-0.00894\\
    14&	Anger&	0	&-0.11058\\
    15&	Sleep problem&	1	&0.01062\\
    16&	Social isolation&	1	&0.2493\\
    17&	Depression&	0	&-0.09246\\
    18&	Humiliated&	0	&-0.02779\\

    \bottomrule
    \end{tabular}%
    \end{adjustbox}
  \label{SAMPLE}%
\end{table}%
In Table \ref{SAMPLE}, we selected a random sample from the dataset and calculated its SHAP values. Figure \ref{shap} illustrates the visualization of Table \ref{SAMPLE}. Features with red colour indicate a positive contribution to the final decision, while features with blue colour indicate negative contributions to the result. Our XGBoost model predicts that this patient has suicide risk in this sample based on his age (56 years), past self-injury experiences, and social isolation problems (which are the most significant positive factors for this patient). Factors that prevent the XGBoost model from identifying this patient has suicidal potential include: not having an anger problem or depression problem, being a Christian (the most important negative factor for this patient), being a widow, and being a clerical worker. The patient was predicted to be at risk for suicide due to stronger positive factors than negative factors. Note that higher positive values in the output indicate a higher risk of committing suicide.

\begin{figure}[!t]
	\centering
	\begin{adjustbox}{width=\linewidth}
		\begin{tabular}{cc}
			\includegraphics[width=0.5\linewidth]{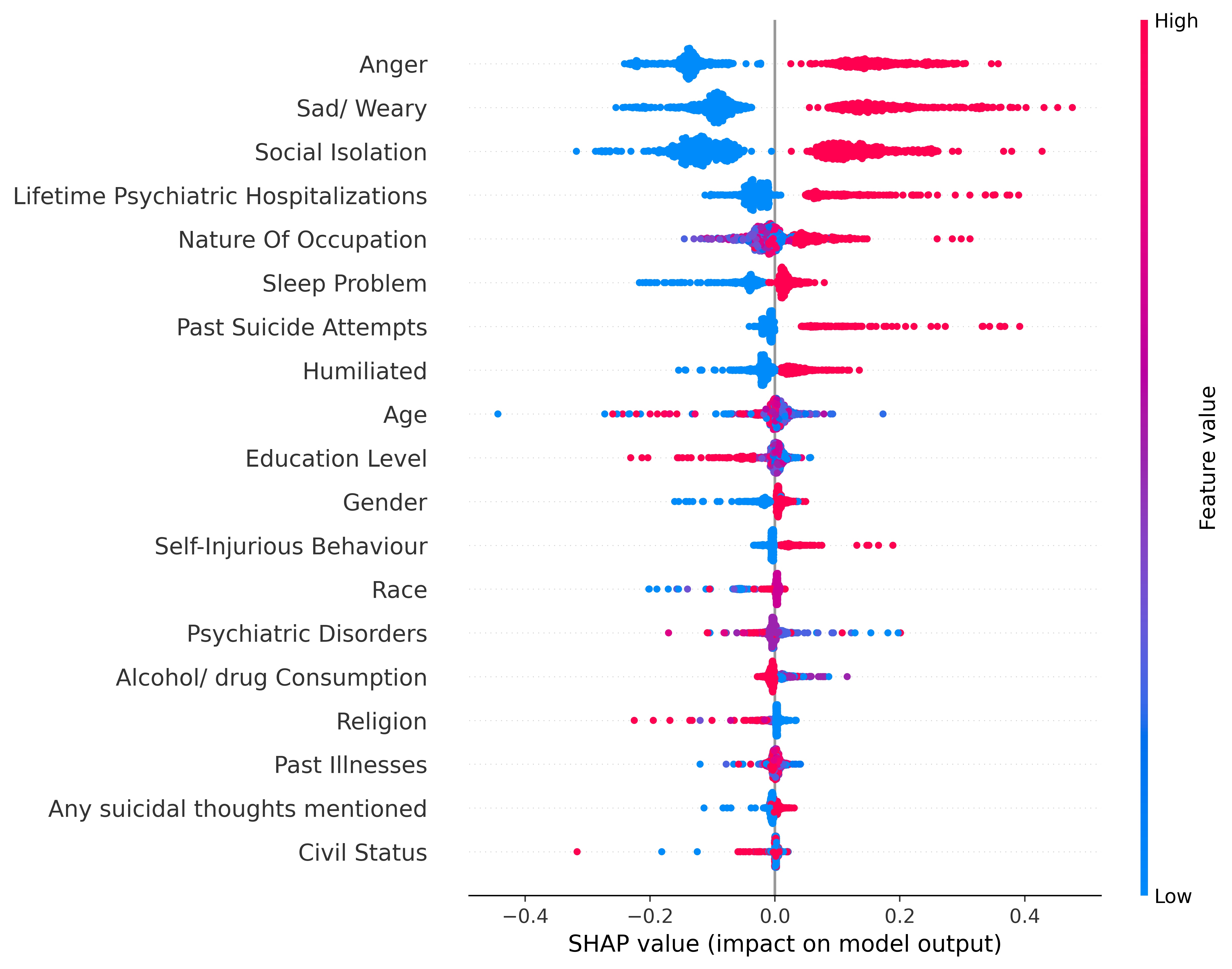} & \includegraphics[width=0.5\linewidth]{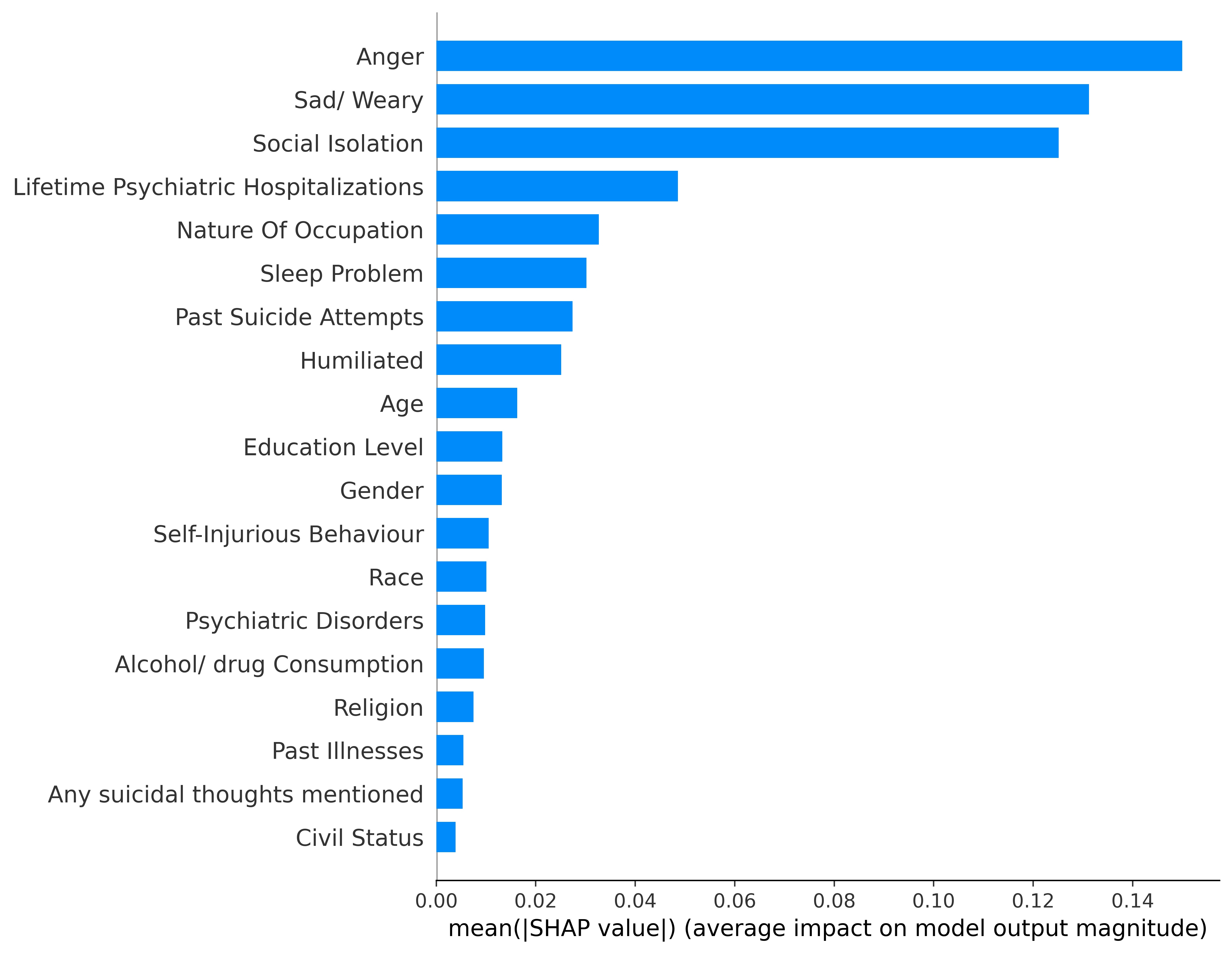}		\\
			(a) & (b)				
		\end{tabular}
	\end{adjustbox}		
	\caption{(a) Overall SHAP Values in the dataset. For each variable and sample, the contribution is shown by the SHAP value. A higher distinction between red and blue points shows higher importance in risk prediction, (b) Feature importance ranking of SHAP analysis. The top 3 variables are the same as the result shown in Figure \ref{python}.}
	\label{fig:Weights}
\end{figure}
The SHAP method also provides interfaces to visualize the overall feature contributions. Figure \ref{fig:Weights} (a) illustrates the overall SHAP value of features in our dataset. Each patient is represented by a point. The red colour points indicate larger feature values, while the bluer colour points indicate lower feature values. It is noteworthy that for features, such as past suicide behaviours and self-injury behaviours, when the values of these features are low, indicating that patients have few related experiences, these variables do not negatively impact the predicted value. However, when the values of these features are high, indicating that these samples have suicidal attempts or self-injuries, these two features significantly contribute to positive prediction.

\begin{figure}[!t] \centering
		\includegraphics[width=0.55\linewidth]{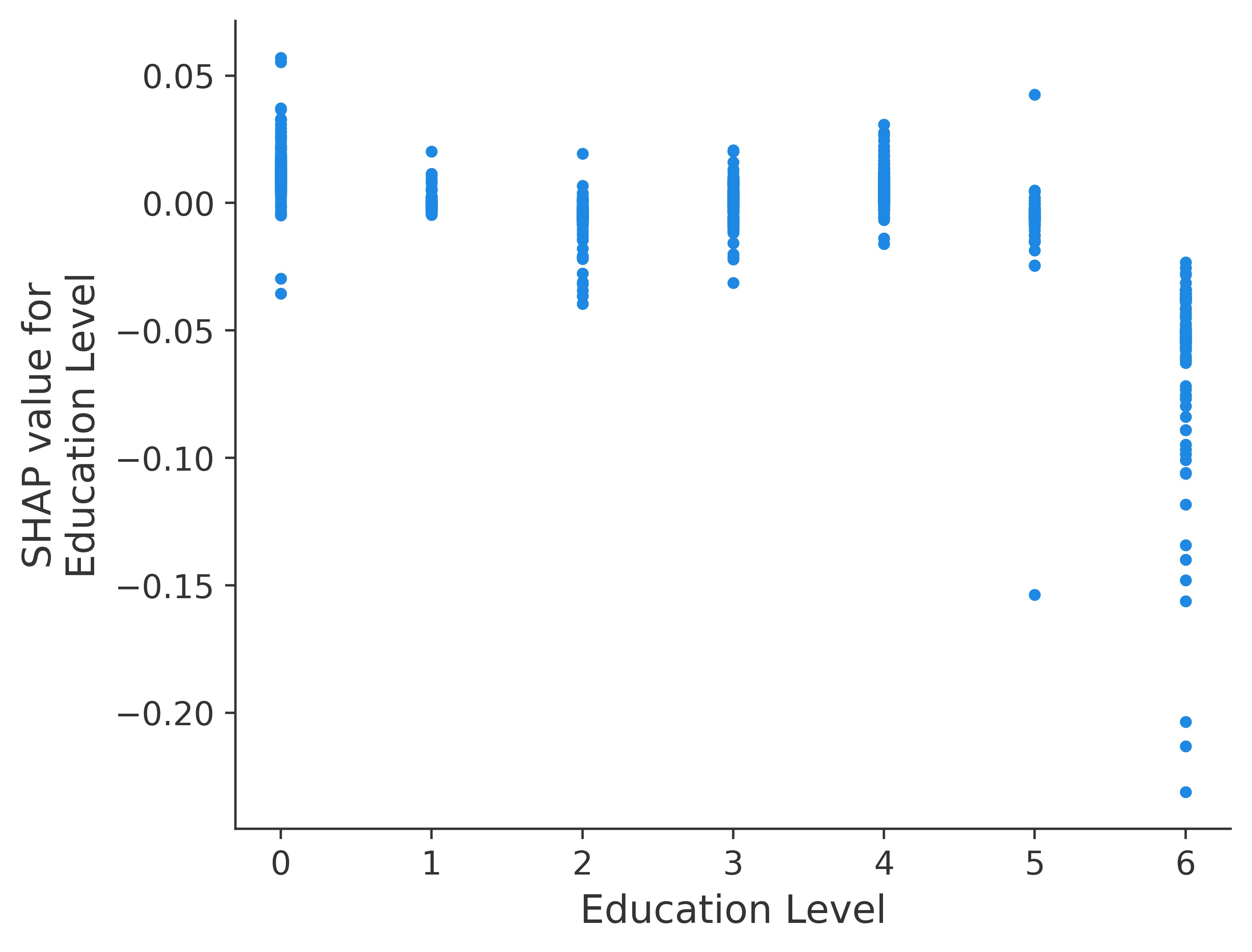}
		\caption{SHAP Value for Educational Level.}
	\label{partial}
\end{figure}

Figure \ref{fig:Weights} (b)  shows the feature importance as calculated by the SHAP package. Although there are some differences compared to Figure \ref{python}, the top-three variables, namely anger problem, depression problem, and social isolation remain the same. According to the importance rank provided by SHAP, psychiatric hospitalization, occupation, and sleeping problem are also crucial features in predicting suicide.
To further analyze the impact of different feature values, the partial dependence plots from SHAP were used. For example in Figure \ref{partial}, each point represents one sample with a corresponding attribute value. It is observed that their distributions are closer to zero for most education levels and tend to be symmetric, indicating that these features do not have a significant  contribution to the final result. Figure \ref{partial} reveals that feature contributions are more pronounced for patients with education level zero (from grade one to seven) and level six (university degree or above). It can be observed that for most patients with education level zero, their SHAP values are positive, indicating a higher risk of suicide, while all patients with education level six have negative SHAP values, indicating a relatively low risk. This suggests that patients with lower levels of education have higher suicide risks, while those with university degrees have lower risks.
\subsection*{Clinical Implications and Future Directions}
Suicide prediction is difficult and traditional self-report-based actuarial risk assessment tools
have been found to have limitations in predicting suicide. The other commonly used methods are clinical judgment and structured professional judgment. Clinical judgment alone has been found to have a sensitivity of $< 25\%$ \cite{appleby2019national}. Most clinicians use structured professional judgment to determine risk. However, there is a need for tools or systems to validate the decision-making in suicide risk prediction. Carter \textit{et al.} \cite{carter2017predicting} undertook a meta-analysis of three types of instruments used to predict suicide death or self-harm: psychological scales, biological tests, and ``third-generation" scales derived from statistical models. This review concluded that no instrument was sufficiently accurate to determine intervention. Similar to other areas of medicine, risk stratification is essential for accurate and precise treatment. The current paper has presented a methodology for improving suicide risk prediction using ML algorithms, which will hopefully increase the confidence of mental health professionals in utilizing ML algorithms in conjunction with clinical risk assessment to improve suicide risk prediction and intervention, and ultimately, to help reverse the trend of increasing suicides worldwide. 
The next step of this study will be to develop a risk assessment interface that utilizes the identified factors and ML algorithm to provide clinicians with a predicted suicide risk for individual patients. This objective risk determination will enhance and refine clinical decision-making and further train the developed models. In future studies, it would be beneficial to investigate other modalities such as speech, image, and videos, as the current ML methods are often trained on text or tabular data. 
\section*{Discussion}
This section justifies the excellent performance of ML algorithms by providing relevant academic evidence to support our experimental results. Firstly, the typical medical diagnosis dataset includes both numerical variables (such as age, past suicide attempts, and blood pressure) and categorical variables (such as gender, marital status, alcohol consumption, sleeping problems, and humiliating experience). This type of data format is ideal for implementing tree-based algorithms. DTs are the most basic tree structure, which classifies each record by evaluating attributes at each node. RF is an advanced version of Decision Trees, which utilizes the Bagging technique to combine the results of multiple trees resulting in improved predictive accuracy. It is observed that the DT model consistently exhibits the best performance when using ML for medical predictions. 

The superior performance of RF algorithms in the healthcare domain has been well-documented in several studies \cite{rajini2021lung,bozorgmehr2021chronic,li2018cardiovascular,raghavendra2020performance}. Research has found that RF models significantly outperform other algorithms in predicting chronic stress and cardiovascular disease risk, with higher accuracy even when using fewer feature variables. The outstanding performance of RF models improves the reliability of diagnosis and reduces the number of tests required for  patients in hospitals. The results and observations made in this paper align with the existing knowledge in  psychiatry and provide a data-driven perspective to justify it. The 50K synthetic+real data analyzed in this paper makes the results general and reliable. The most important variables identified in this study can serve as a foundation for future research in the field.  

\begin{figure*} [!t]\centering
		\includegraphics[width=1\linewidth]{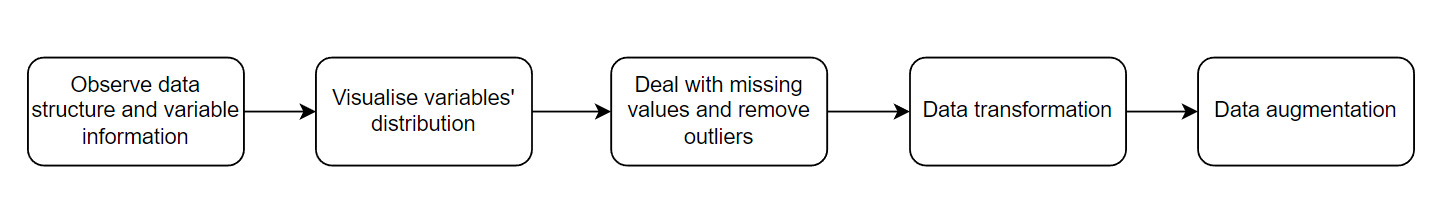}
		\caption{Data preparation steps.}
	\label{flow}
\end{figure*}

The current paper aims to evaluate the performance of Machine Learning (ML) algorithms in predicting suicide and to improve the interpretability of ML models by using XAI models. To achieve these objectives, the paper implements the entire process of using ML algorithms to predict suicide with XAI. Firstly, we conducted a literature review to summarise state-of-the-art suicidal datasets, psychometric questionnaires, ML models, and model evaluation parameters. Secondly, to prevent under-fitting when building models, the CTGAN and Scikit-learn are used to generate an artificial dataset. The CTGAN method had many powerful functions for data augmentation, but in terms of the distribution of feature values, the dataset generated by the Scikit-learn method more closely resembles the distribution of the original dataset. In this paper, seven models were built and repeated experiments were conducted to evaluate their performance. The Decision Tree (DT), Random Forrest (RF), and XGBoost models all showed excellent performance among the seven models. Correlation analysis revealed that mental health problems are strongly related to suicidal behaviours, which is consistent with existing research findings. Additionally, the XAI framework was applied to identify the dominant and key factors associated with suicide, which included anger problems, depression problems, social isolation, psychiatric hospitalization, and patients’ occupation. 
\section*{Method}
\subsection*{Dataset Selection}
To determine the best  dataset to use, we considered three criteria: \textbf{(i)} it should have a sufficient number of variables related to mental issues, \textbf{(ii)} the key outcome  should be labeled indicating whether the patient had committed suicide or not, \textbf{(iii)} it should be sufficiently large-scale to successfully apply ML techniques. 
In addition to these main objectives, the dataset variables should be easy to understand and should include textual variables, such as interview excerpts and symptom descriptions, to allow for  the use of NLP techniques.  The word cloud method developed by \cite{cook2016novel} can be used to effectively extract keywords from massive text information and intuitively visualize the importance of keywords. Based on above criteria, we selected our dataset (\href{https://github.com/dinisurunisal/Suicide-Risk-Prediction-Project/blob/master/DataScience/AlgorithmComparison/Test-Data-10.csv}{link}) from a survey in England that was publicly released for university student research. The raw dataset contained 22 variables and approximately 1K records. Briefly, this dataset includes 12 attributes (as shown in Table \ref{SAMPLE}) indicating patients' mental illness problems, including past suicide attempts, suicidal thoughts, self-injuries, psychiatric disorders, and more. Additionally, some numerical and categorical variables provided comprehensive information about the patients, such as their occupation, marriage status, and education level, which are valuable in a prediction task.
\subsection*{Data Preparation}
Data-related tasks and analyses were implemented using the Python programming language and its open-source libraries. 
Figure \ref{flow} illustrates the steps taken for the data preparation in this paper.

\begin{table*}[t!]
\caption{The mean and standard deviation of variables in the original and the augmented datasets in the form of (m,s).}
  \centering
 \begin{adjustbox}{scale=0.85,angle=0}
    \begin{tabular}{l|ll|ll|ll}

    \hline
   \textbf{Dataset$\rightarrow$} & \multicolumn{2}{l|}{\textbf{Original}} &  \multicolumn{2}{l|}{\textbf{Scikit-learn}} & \multicolumn{2}{l}{\textbf{CTGAN}}\\ 
       \hline 
   \textbf{Features $\downarrow$/Group by$\rightarrow$}& Suicide & Not Suicide & Suicide & Not Suicide & Suicide & Not Suicide \\
    \hline
      Patient age & (49.01,20.98) & (54.63,23.01) & (51.04,17.38) & (52.96,16.99) & (59.21, 20.54) & (58.61,20.92)\\
    Past suicide attempts&(0.29,0.46)&(0.05,0.22)&(0.28,0.45)&(0.11,0.31)&(0.11,0.31)&(0.11,0.31)\\
    Suicide thoughts&(0.54,0.5)&	(0.08,0.27)&	(0.5,0.5)&	(0.09,0.29)&	(0.07,0.25)&	(0.06,0.23)\\
    Self-injuries&	(0.40,0.49)&	(0.05,0.22)&	(0.31,0.46)&	(0.11,0.31)&	(0.04,0.21)&	(0.05,0.23)\\
    Anger problem&	(0.82,0.39)&	(0.1,0.31)&	(0.58,0.49)&	(0.21,0.41)&	(0.24,0.43)&	(0.22,0.41)\\
    Sleeping problem&	(0.88,0.33)&	(0.37,0.48)&	(0.68,0.47)&	(0.55,0.5)&	(0.35,0.48)&	(0.32,0.46)\\
    Social isolation&	(0.77,0.42)&	(0.08,0.27)&	(0.64,0.48)&	(0.09,0.29)&	(0.09,0.28)&	(0.1,0.3)\\
   Depression problem&	(0.74,0.44)&	(0.06,0.24)	&(0.56,0.5)&	(0.02,0.14)&	(0.16,0.37)&	(0.14,0.35)\\
   Humiliated experience&	(0.67,0.47)&	(0.05,0.22)&	(0.63.0.48)&	(0.10,0.3)&	(0.09,0.29)&	(0.09,0.29)\\
    \bottomrule
    \end{tabular}%
    \end{adjustbox}
  \label{dist}%
\end{table*}%

\subsection*{Data Augmentation}
The dataset only includes 1,000 records which are not sufficient to train a robust and high-performance deep learning model. To increase the dataset size, we implemented two data augmentation methods to enhance the dataset. One is from the Conditional Generative Adversarial Network (CTGAN) \cite{lee2021ctgan} and the other is from the pre-processing module of the Scikit-learn package.\\
The Tabular Generative Adversarial Network (TGAN) is the initial version of the CTGAN method. TGAN utilizes synthetic data generated by the conditional generative adversarial networks and has shown better performance than existing deep learning methods \cite{xu2019modeling}. Moreover, CTGAN has several advanced functions, such as setting boundaries when generating numerical variables, conditional sampling, and creating primary keys for the dataset, which are added benefits. \\
On the other hand, the module used from the Scikit-learn package ensures that there is a similarity in the distributions between the original data and the data which is synthetically generated using encoders.
Experiments were conducted to evaluate the performance of these two techniques on the selected dataset. We augmented the original data to 50,000 samples and recorded the distributional information of each variable. To avoid randomness, we repeated this process 10 times to obtain reliable averaged results. Table \ref{dist} illustrates the first and second moments of variables' distributions in the original dataset and the synthetic datasets. After encoding variables, the program mapped the values of psychiatric variables into two categories, where zero indicates that the patient does not have this problem, and one indicates the patient does. Taking a humiliating experience as an example, a value of one means that the patient has been humiliated in the past, while a value of zero means that the patient has not experienced humiliation. 
Among these variables, the statistics of Scikit-learn generated data are much closer to the original data mean, which indicates that the tools in Scikit-learn maintain the original distribution better than CTGAN. Therefore, we decided to choose the synthetic dataset generated by Scikit-learn as the augmented data for further steps.

Notably, our main goal of implementing data augmentation is to prevent under-fitting. However, an excessive dataset size may result in poor accuracy of the model. Therefore, a proper compromise is achieved by using a sample size of 50,000.

\bibliography{reference}
\section*{Data Availability}
The dataset used for this study is a publicly available dataset which can be found in \url{https://github.com/dinisurunisal/Suicide-Risk-Prediction-Project/blob/master/DataScience/AlgorithmComparison/Test-Data-10.csv}  
\section*{Code Availability}
The code for training and testing of machine learning models used in this study will be available on GitHub, subject to its acceptance.
\section*{Competing Interests}
The  authors  declared  no  potential  conflicts  of  interest  with  respect  to  the  research,  authorship  and/or  publication  of  this  article.
\section*{Author Contribution}
M.B., D.R., G.D. and F.S. conceived of the study, acquired funding, and were responsible for the
overall study; H.T and A.M.R. conducted the analyses; H.T. and A.M.R.
drafted the manuscript; all other authors critically reviewed and commented on the
analyses and manuscript.
\section*{Ethics Approval}
This project was approved by the Royal Perth Hospital Human Research Ethics Committee (approval number RGS 4360). No personal data was processed in this study and the dataset used for this study is a publicly available anonymized dataset.
\section*{Funding}
This project is funded by EMHS Mental Health Research.

\end{document}